\documentclass[10pt,twocolumn,letterpaper]{article}

\usepackage{icb}
\usepackage{times}
\usepackage{epsfig}
\usepackage{graphicx}
\usepackage{amsmath}
\usepackage{amssymb}
\usepackage{multirow}
\usepackage{subfigure}
\usepackage[breaklinks=true]{hyperref}

\icbfinalcopy 

\ificbfinal\fi
\begin{document}

\title{Periocular Recognition in the Wild with Orthogonal Combination of Local Binary Coded Pattern in Dual-stream Convolutional Neural Network}

\author{Leslie Ching Ow Tiong\\
KAIST\\
291 Daehak-ro, Yuseong-gu, \\
Daejeon 34141, \\
Republic of Korea\\
{\tt\small tiongleslie@kaist.ac.kr}
\and
Andrew Beng Jin Teoh\\
Yonsei University\\
50 Yonsei-ro, Sinchon-dong,\\
Seodaemun-gu, Seoul,\\
Republic of Korea\\
{\tt\small bjteoh@yonsei.ac.kr}
\and
Yunli Lee\\
Sunway University\\
5 Jln. Universiti, Bdr. Sunway, \\47500 Petaling Jaya, Selangor,\\ Malaysia\\
{\tt\small yunlil@sunway.edu.my}}
\maketitle
\thispagestyle{empty}

\begin{abstract}
In spite of the advancements made in the periocular recognition, the dataset and periocular recognition in the wild remains a challenge. In this paper, we propose a multilayer fusion approach by means of a pair of shared parameters (dual-stream) convolutional neural network where each network accepts RGB data and a novel colour-based texture descriptor, namely Orthogonal Combination-Local Binary Coded Pattern (OC-LBCP) for periocular recognition in the wild. Specifically, two distinct late-fusion layers are introduced in the dual-stream network to aggregate the RGB data and OC-LBCP. Thus, the network beneficial from this new feature of the late-fusion layers for accuracy performance gain. We also introduce and share a new dataset for periocular in the wild, namely Ethnic-ocular dataset for benchmarking. The proposed network has also been assessed on one publicly available dataset, namely UBIPr. The proposed network outperforms several competing approaches on these dataset.
\end{abstract}

\section{Introduction}
In recent years, periocular recognition is gaining attention from the biometrics community due to its promising recognition performance \cite{Barroso2016}. Periocular usually refers to the region around the eyes, preferably including the eyebrow. An early study of the periocular recognition was done by Park \etal \cite{Park2009}, which demonstrated promising results in controlled environments. The authors utilised several texture descriptors such as Histogram of Orientation and Gradient (HOG), Local Binary Pattern (LBP), and Scale Invariant Feature Transform (SIFT), followed by score fusion for decision. Several studies such as \cite{Mahalingam2013} also focus on using texture descriptors and learning models for periocular recognition. They combined several texture descriptors for a better feature representation. Another work reported in \cite{Cao2016}, convolved the HOG and LBP that generated from periocular images with Gabor filters and followed by concatenation. Although all the existing approaches achieved decent recognition performances, these approaches were less robust to the ``in the wild'' variations such as pose alignments, illuminations, glasses, and occlusions.

Since 2012, Convolutional Neural Network (CNN) has gained an exponential attention to learn high-dimensional data in the computer vision domain \cite{Krizhevsky2012}. CNN with colour-based texture descriptors have successfully been employed in numerous vision applications, such as emotional recognition \cite{Levi2015} and texture classification \cite{Anwer2018}. Both \cite{Levi2015} and \cite{Anwer2018} demonstrated that using the colour texture do provide complementary information to improve CNN in extracting feature representations. In their analysis, they showed that texture descriptors are ubiquitous enough to represent an object, especially when the shape cannot be visualised clearly.

For periocular recognition in controlled environments, \cite{Gangwar2016} exploited two CNNs, which extract comprehensive periocular information from left and right oculars. \cite{Proenca2018} and \cite{Zhao2018} focused on the feature extraction based on regions-of-interest of periocular with CNNs. Both networks exploit prior knowledge by discarding unnecessary information to enhance CNN in periocular recognition. However, these networks are not well-performed in the wild environment, such as when the periocular images are misaligned or the periocular images does not include the eyebrows or ocular perfectly. A very recent work \cite{Soleymani2018} proposed a multimodal CNN, namely multi-abstract fusion CNN where the features fusion for iris, face and fingerprint takes place at fully-connected (\textit{fc}) layer. A fusion layer is designed to fuse the different levels of \textit{fc} layers as multi-feature representations with sole RGB data. However, this work only limited to RGB data, which could be of limited in information.

In this paper, we investigate a fusion approach with a dual-stream CNN where each network accepts RGB data and a novel colour-based texture descriptor, namely Orthogonal Combination-Local Binary Coded Pattern (OC-LBCP) for periocular recognition in the wild. Both networks are shared in parameters and a late fusion takes place at the last convolutional (\textit{conv}) layer before \textit{fc} layer. OC-LBCP exploits the colour information in texture representation and can extract features with higher discriminative capability.

This paper also attempts to address periocular recognition in the wild challenge, which remains not well-covered by the existing datasets \cite{CASIA, Padole2012, Raghavendra2016} and research community \cite{Gangwar2016, Soleymani2018}. The periocular recognition in the wild challenges are associated to the huge differences of the periocular images due to sensors location, pose alignments, level of illuminations, occlusions, and others. Specifically, the appearances of periocular region images with cosmetic products and plastic surgery may jeopardise the accuracy performance severely. Most of the existing periocular datasets such as CASIA-iris distance \cite{CASIA} dataset and UBIPr dataset \cite{Padole2012} were collected randomly under controlled environments and contained limitation for ethnicity. \cite{Rhee2012} also revealed that each ethnic group has a unique shape of periocular and skin texture of periocular regions.

We therefore create a new dataset, namely Ethnic-ocular\footnote {Ethnic-ocular dataset is available at: \url{https://www.dropbox.com/sh/vgg709to25o01or/AAB4-20q0nXYmgDPTYdBejg0a?dl=0}} by collecting the periocular region images in the wild to validate the proposed method. The dataset is created based on five ethnic groups namely \textit{African}, \textit{Asian}, \textit{Latin American}, \textit{Middle Eastern}, and \textit{White}. As a result, our dataset is designed in such way to avoid unbalanced selection as there are differences in the configuration of oculars among different ethnicities.

Thus, the contributions of this paper are as follows:
\vspace{-7pt}
\begin{itemize}
\setlength{\itemsep}{0pt}
\setlength{\parskip}{0pt}
\setlength{\parsep}{0pt}
\item Investigate periocular recognition in the wild with the combination of RGB data and the proposed colour-based texture descriptor OC-LBCP for dual-stream CNN.
\item To offer a better feature representation for periocular recognition in the wild, two distinct late-fusion layers are introduced in the dual-stream CNN. The role of the late-fusion layers is to aggregate the RGB data and OC-LBCP. Thus, the dual-stream CNN beneficial from this new features of the late-fusion layers to deliver better accuracy performance.
\item A new periocular dataset, namely Ethnic-ocular dataset is created, containing periocular images in the wild. The images are collected across highly uncontrolled subject-camera distances, low-resolution images, different appearances, occlusions, glasses, poses, time, locations, and illuminations. The dataset also provides training and testing schemes for performance analysis and evaluation.
\end{itemize}

This paper is organised as follows; Section 2 presents the structure of OC-LBCP. Section 3 explains the presentation of our proposed network, dual-stream CNN with late-fusion layers. Section 4 presents the detailed information of proposed dataset. Section 5 describes the presentation of experimental analysis and results. A conclusion is summarised in Section 6.

\section{Colour-based texture descriptor}
\begin{figure}[t]
\begin{center}
   \includegraphics[width=0.83\linewidth]{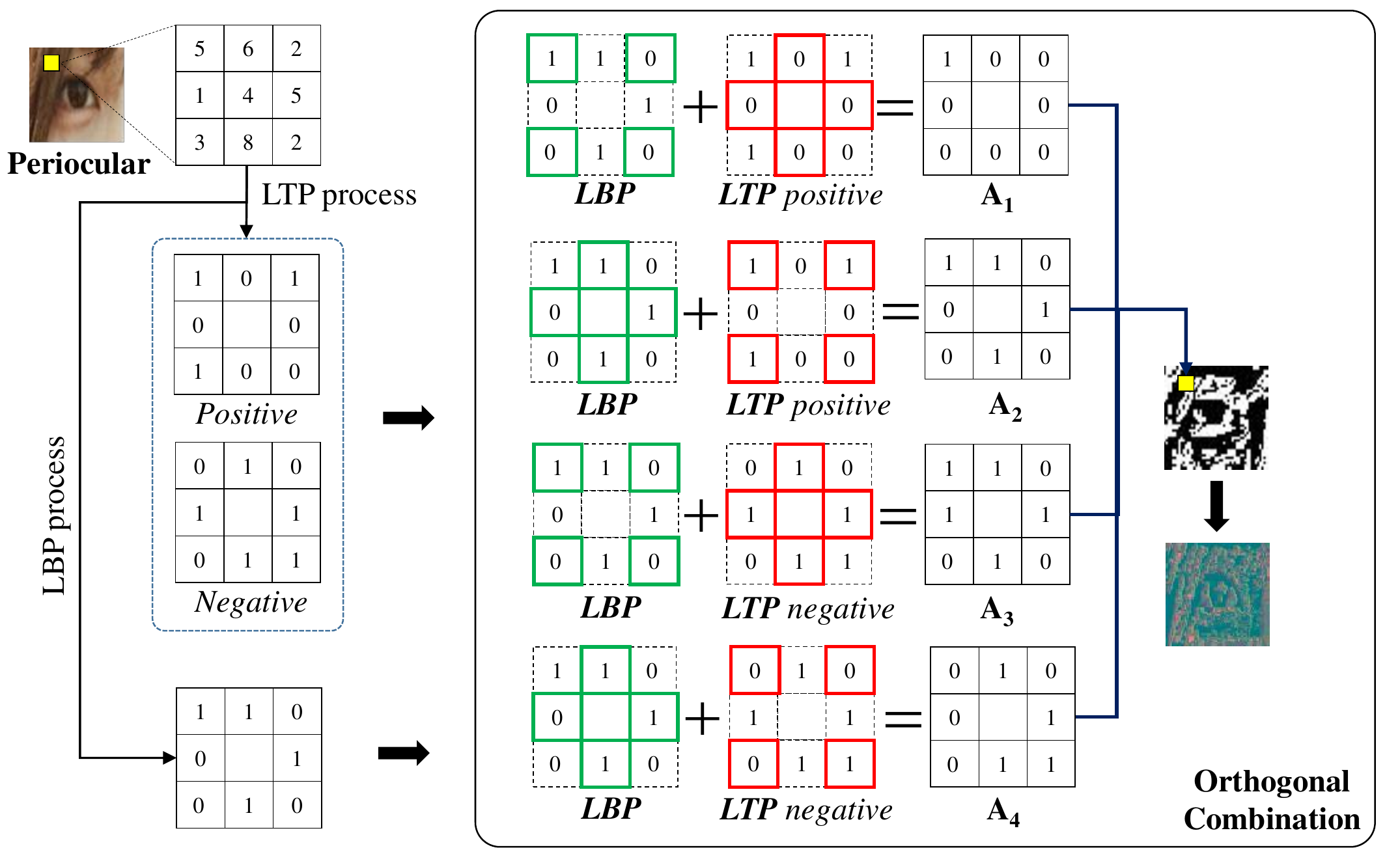}
\end{center}
\vspace{-10pt}
   \caption{Illustration of OC-LBCP.}
\label{fig:fig1}
\vspace{-10pt}
\end{figure}

We devise a colour-based texture descriptor known as OC-LBCP by means of orthogonal combination of Local Binaray Pattern (LBP) \cite{Ojala2002} and Local Ternary Pattern (LTP) \cite{Tan2010}. LBP is to summarise the local structure in an image by comparing each pixel with its neighbourhood. This descriptor works by thresholding a 3$\times$3 neighbourhood using the grey level of the central pixel in the binary code. LTP is an extension of the primary LBP with three-valued codes by neighbouring pixels with thresholding to form a ternary pattern. The ternary pattern results in a large range, so the ternary pattern is split into two positive and negative binary patterns as depicted in Figure \ref{fig:fig1}.

The OC-LBCP is designed to reduce the sensitivity of image noise and levels of illumination by creating better texture information of an object. Suppose $\textbf{I} \in \mathbb{R}^{h \times w}$ be an image, where $h$ and $w$ are the height and width, respectively. Butterworth filter is applied to $\textbf{I}$ to separate the illumination component from $\textbf{I}$ and enhance the reflectance \cite{Delac2006}.

Next, orthogonal combination of binary codes from LBP and LTP operators is carried out. Figure \ref{fig:fig1} illustrates the orthogonal combination by concatenating the LBP and LTP binary codes into four orthogonal groups, namely $A_{1}$, $A_{2}$, $A_{3}$, and $A_{4}$. The orthogonal combinations are beneficial to achieve illumination invariance by removing outlying disturbances. To generate $A_{1}$, we select the bits from red line boxes in LTP positive and the values from green line boxes in LBP (see Figure \ref{fig:fig1}); then, concatenated all of them. The same process is repeated for $A_{2}$, $A_{3}$, and $A_{4}$. The OC-LBCP $\Omega(i,j)$ is formed by choosing the largest binary codes from the orthogonal groups. The OC-LBCP is formed by combining the features as follows:
\begin{align}
    \Omega(i,j) &= \textrm{max}(A(i,j)), \\
    A(i,j) &= \{ A_{1}(i,j), A_{2}(i,j), A_{3}(i,j), A_{4}(i,j) \},
    \label{eq:eq1}
\end{align}
where $A$ represents the four orthogonal groups with binary codes.

To map $\Omega(i,j)$ into a colour space, we define a colour pattern matrix $\Delta (u,v)$ to represent the similarity of the image intensity patterns across all possible code values based on \cite{Levi2015}. The colour mapping transforms OC-LBCP into a colour-based texture representation, which reflects the differences in the intensity patterns.

The colour pattern matrix is computed based on the Earth Movers Distance by following \cite{Levi2015}. Then, apply Multi-Dimensional Scaling (MDS) to seek a mapping of the codes into a low dimensional metric space:
\begin{equation}
\Delta(u,v) = \left \lfloor \textrm{MDS}[\Omega_{u}] + \textrm{MDS}[\Omega_{v}] \right \rfloor
\end{equation}
where $u$ and $v$ are the coordinates in colour pattern matrix $\Delta(u, v)$ and $\left \lfloor \cdot \right \rfloor$ is the floor function.

\section{Dual-stream convolutional neural network}
\begin{figure}[t]
\begin{center}
   \includegraphics[width=0.85\linewidth]{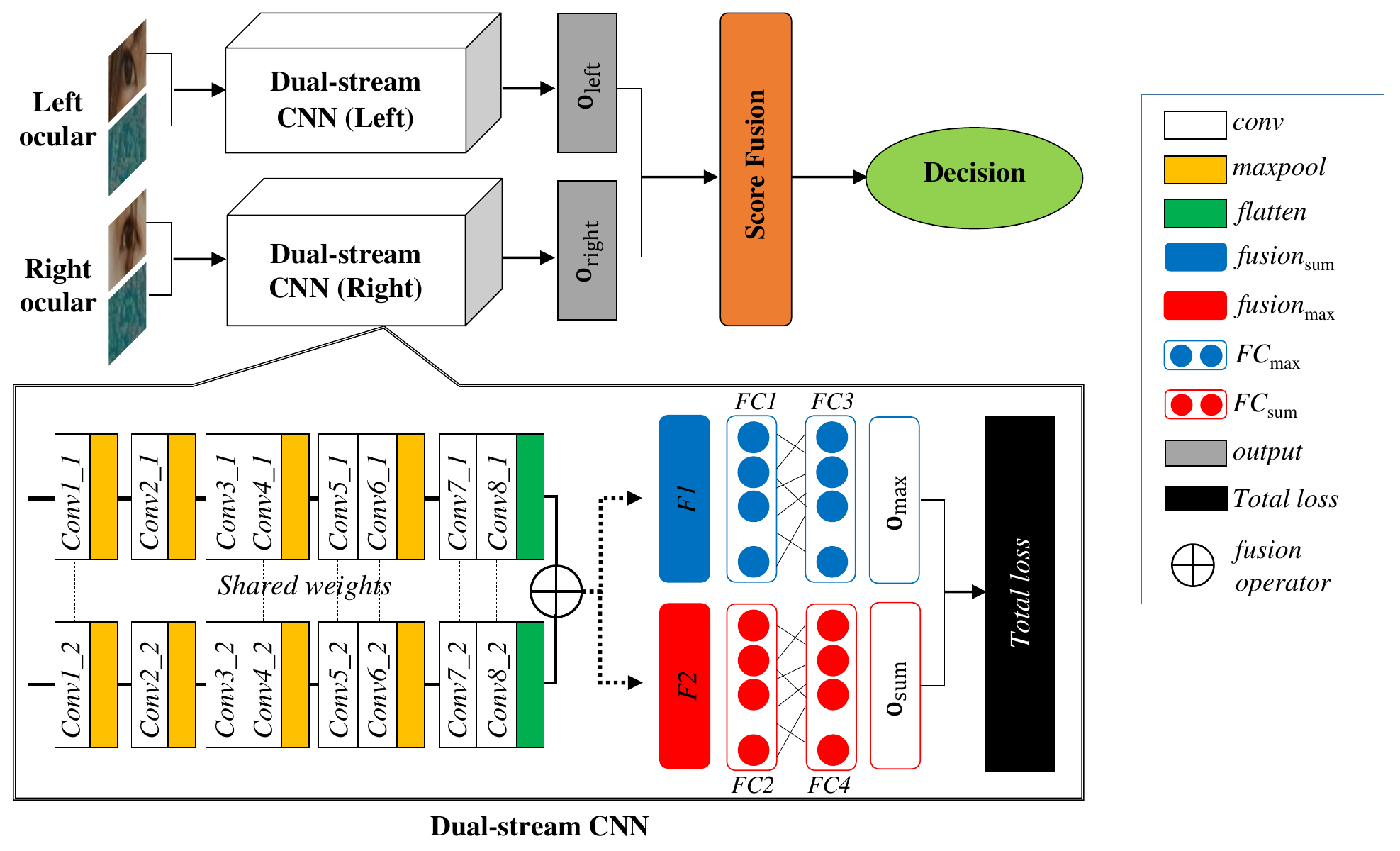}
\end{center}
\vspace{-10pt}
   \caption{Architecture of our proposed network.}
\label{fig:fig2}
\end{figure}

Dual-stream CNN was originally designed to extract features from temporal and structural streams for action detection and recognition \cite{Feichtenhofer2016}. Our work is motivated by the dual-stream CNN where RGB data and texture descriptor are conceived as the first and second stream. As shown in Figure \ref{fig:fig2}, the network consists of eight pairs of shared \textit{conv} layers and eight max-pooling (\textit{maxpool}) layers, where it is designed to learn the correspondence between the RGB data and OC-LBCP and to discriminate between themselves with the shared weights. Table \ref{tab:tab1} tabulates the proposed network architecture. The shared \textit{conv} layer is given as a pair \textit{conv} layer, with their parameters shared.

\subsection{Late-fusion layers}
To integrate the information from the dual inputs, we merge the flatten layers ($F1$ and $F2$ in Table \ref{tab:tab1}) by creating late-fusion layers to strengthen the feature activations of the network. Therefore, two fusion layers, namely \textit{max} and \textit{sum} layers are introduced, to fuse the features at flatten layers. To be specific, \textit{max} layer $Z_{\textrm{max}}$ takes a larger activation from $F1$ or $F2$ with $N$ nodes ($N=12,800$). On the other hand, \textit{sum} fusion layer $Z_{\textrm{sum}}$ takes a sum of activations of $F1$ and $F2$.

\subsection{Training with total loss}
For training, a total loss function that composed of summation of cross entropy of logit vector of \textit{max} fusion and \textit{sum} fusion and their respective one-hot encoded labels as follows is utilised:
\begin{align}
    total_{\textrm{loss}} &= \mathcal{L} \left ( \textbf{y}_{\textrm{max}} \right ) + \mathcal{L} \left ( \textbf{y}_{\textrm{sum}} \right ), \\
    \mathcal{L}(\textbf{y}_{*}) &= - \sum_{i}^{M} \sum_{j}^{C} l_{ij} \textrm{log} ( \textrm{softmax}(\textbf{y}_{*})_{ij}),
    \label{eq:eq2}
\end{align}
where $* \in \{ \textrm{max}, \textrm{sum} \}$. $l$, $M$ and $C$ denote class label, the numbers of training sample, and the number of class, respectively. Since a periocular region can be either left or right oculars; we therefore train each side with separate dual-stream CNN (see Figure \ref{fig:fig2}).

\begin{table}[!t]
\centering
\fontsize{8.5}{9.5}\selectfont
\caption{Configuration of the proposed network. Note that, \textit{f.m.} is defined as the size of output $conv$ layer and \textit{f.} as the size of filter.}
\label{tab:tab1}
\begin{tabular}{|l|c|}
\hline
\multicolumn{1}{|c|}{\textbf{Layer}} & \textbf{Configuration} \\ \hline
$conv1\_1$, $conv1\_2$ & \textit{f.m.}: 64$\times$80$\times$80; \textit{f.}: 2$\times$2; maxpool: 2$\times$2\\ \hline
$conv2\_1$, $conv2\_2$ & \textit{f.m.}: 128$\times$40$\times$40; \textit{f.}: 2$\times$2; maxpool: 2$\times$2\\ \hline
$conv3\_1$, $conv3\_2$ & \textit{f.m.}: 256$\times$20$\times$20; \textit{f.}: 2$\times$2\\ \hline
$conv4\_1$, $conv4\_2$ & \textit{f.m.}: 256$\times$20$\times$20; \textit{f.}: 2$\times$2; maxpool: 2$\times$2\\ \hline
$conv5\_1$, $conv5\_2$ & \textit{f.m.}: 512$\times$10$\times$10; \textit{f.}: 2$\times$2\\ \hline
$conv6\_1$, $conv6\_2$ & \textit{f.m.}: 512$\times$10$\times$10; \textit{f.}: 2$\times$2; maxpool: 2$\times$2\\ \hline
$conv7\_1$, $conv7\_2$ & \textit{f.m.}: 512$\times$5$\times$5; \textit{f.}: 2$\times$2\\ \hline
$conv8\_1$, $conv8\_2$ & \textit{f.m.}: 512$\times$5$\times$5; \textit{f.}: 2$\times$2\\ \hline
$F1$, $F2$ & 1$\times$1$\times$4,096 \\ \hline
$Z_{\textnormal{max}}$, $Z_{\textnormal{sum}}$ & 1$\times$1$\times$4,096 \\ \hline
$FC1$, $FC2$ & 1$\times$1$\times$4,096 \\ \hline
$FC3$, $FC4$ & 1$\times$1$\times$4,096 \\ \hline
$\textbf{o}_{\textrm{max}}$, $\textbf{o}_{\textrm{sum}}$ & 1$\times$1$\times C$ \\ \hline
\end{tabular}
\vspace{-10pt}
\end{table}

\subsection{Testing with score fusion layer}
Let $\textbf{o}_{\textrm{max}}= \textrm{softmax}(\textbf{y}_{\textrm{max}})\in \mathbb{R}^{C}$ and $\textbf{o}_{\textrm{sum}}= \textrm{softmax}(\textbf{y}_{\textrm{sum}})\in \mathbb{R}^{C}$ be the softmax vectors of respective \textit{max} (after $FC3$) and \textit{sum} ($FC4$) output layer and $C$ is the number of classes. The two softmax vectors are aggregated yield $\textbf{o} = \textbf{o}_{\textrm{max}} + \textbf{o}_{\textrm{sum}} \in \mathbb{R}^{C}$. Since there are two dual-stream CNN, each for left and right ocular, thus we differentiate the softmax vector $\textbf{o}$ to $\textbf{o}_{\textrm{left}}$ and $\textbf{o}_{\textrm{right}}$, respectively.

To determine the identity of an unknown input based on the trained network, we follow the identification protocol where the testing set, which is not overlapping with training set is divided according to gallery and probe sets. Each subject in the gallery set is composed of his/her left and right softmax vectors $\textbf{o}_{i}^{G} = \{ \textbf{o}_{\textrm{left},i}^{G}, \textbf{o}_{\textrm{right},i}^{G} \}$ where $i=1, \cdots, C$.

For a given probe with its left and right ocular softmax vectors $\textbf{o}^{P} = \{ \textbf{o}_{\textrm{left}}^{P}, \textbf{o}_{\textrm{right}}^{P} \}$, we compute the fused score with sum rule as:
\begin{equation}
s_{\textrm{fuse}} (\textbf{o}_{i}^{G}, \textbf{o}^{P}) = s(\textbf{o}_{\textrm{left},i}^{G}, \textbf{o}_{\textrm{left}}^{P}) + s(\textbf{o}_{\textrm{right},i}^{G}, \textbf{o}_{\textrm{right}}^{P})
\end{equation}
where $s(\textbf{o}_{\textrm{*},i}^{G}, \textbf{o}_{\textrm{*}}^{P}) = 1 - \textrm{cos}(\textbf{o}_{\textrm{*},i}^{G}, \textbf{o}_{\textrm{*}}^{P}), * \in \{ \textrm{left},  \textrm{right} \}$ and $\textrm{cos}(\cdot, \cdot)$ is the cosine similarity distance. Finally, the identity of $\textbf{o}^{P}$, $\delta$ can be decided based on
\begin{equation}
\delta = \underset{i}{\max} [ s_{\textrm{fuse}} (\textbf{o}_{i}^{G}, \textbf{o}^{P}) ].
\end{equation}

\section{Ethnic-ocular dataset}
To design our dataset, we follow the example of FaceScrub \cite{Ng2014} dataset collection. Our goal is to produce a large collection of periocular images based on different ethnic groups for recognising individuals. Thus, all the periocular images are collected in the wild, such as uncontrolled subject-camera distances, locations, poses, appearances with and without make-up, and level of illuminations.

\subsection{Collection setup and information}
We began with 250 subject names from FaceScrub dataset and the 784 subject names from BBC News \cite{BBC}, CNN News \cite{CNN}, and Naver News \cite{Naver}, in order to search for the images of these subjects across Google image search engine. In the search, the top 300 images for each subject were downloaded. We first extracted facial regions in these images by using the Viola-Jones face detector from Matlab \cite{Matlab}. The views of facial region in these images are between -45$^{\circ}$ and 45$^{\circ}$. The images were manually verified to ensure that the images are correctly labelled by the subjects.

\begin{figure}[t]
\begin{center}
   \includegraphics[width=0.45\linewidth]{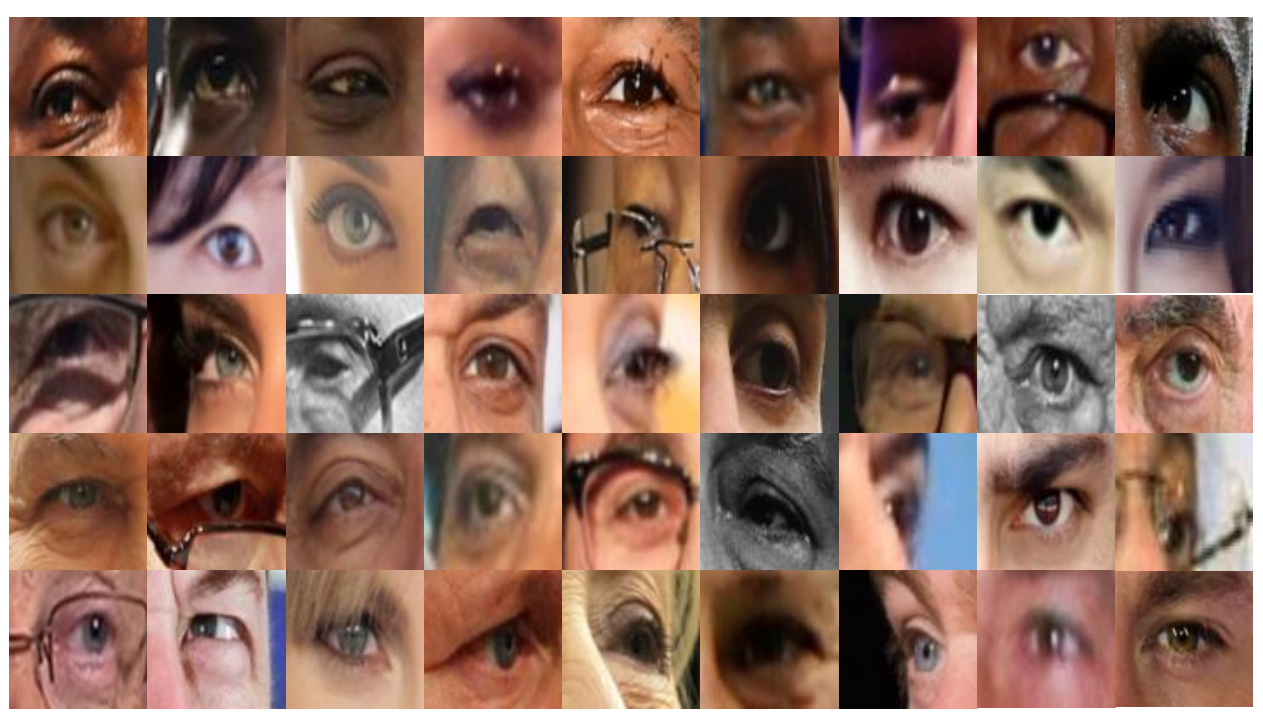}
\end{center}
\vspace{-10pt}
   \caption{Sample images of Ethnic-ocular dataset. Each row
presents different images per ethnic group.}
\label{fig:fig3}
\vspace{-10pt}
\end{figure}

The dataset contains 85,394 images (includes left and right oculars) of 1,034 subjects. To extract the periocular region from each image, we first aligned all the images by fixing the coordinates of facial feature points based on the Viola-Jones face detector bounding box. Then, the images were cropped into left and right oculars by using the technique from \cite{Struc2010}, and the results were resized to 80$\times$80 pixels individually as shown in Figure \ref{fig:fig3}.

\subsection{Benchmark protocols}
The dataset provides training and testing protocols; 623 subjects were randomly selected as training and the rest of the subjects were used as testing. In the testing, we have divided the images such that the ratio between the gallery sets and probe sets is 50:50. This division process was repeated three times.

\section{Experiments}
We used the proposed dataset namely Ethnic-ocular dataset and one public dataset - UBIPr \cite{Padole2012} as the target datasets to evaluate the performances of the dual-stream CNN and other benchmark approaches. All the configurations of approaches are described next.

\subsection{Experimental setup}
\subsubsection{Proposed network}
\hspace{8pt} Dual-stream CNN is implemented by using the TensorFlow \cite{Tensor} toolkit. We applied an annealed learning rate, which started from $1.0 \times 10^{-3}$ and it is subsequently reduced by $10^{-1}$ for every 10 epochs. The minimum learning rate was defined as $1.0 \times 10^{-5}$. An Adam optimizer was applied, where the weight decay and momentum were set to $5.0 \times 10^{-4}$ and 0.9, respectively. The batch size was set to 64 and the training was carried out across 200 epochs. The training was done by using our dataset and it was performed by an NVidia Titan Xp GPU.

\vspace{-10pt}
\subsubsection{Benchmark approaches}
\hspace{8pt} Seven deep networks were selected to evaluate the performance of periocular recognition, namely AlexNet \cite{Krizhevsky2012}, FaceNet \cite{Schroff2015}, LCNN-29 \cite{Wu2018}, VGG-16 \cite{Parkhi2015}, DeepIrisNet-A \cite{Gangwar2016}, DeepIrisNet-B \cite{Gangwar2016}, and Multi-abstract fusion CNN \cite{Soleymani2018}. Here we use the pre-trained models that were provided by the authors. All the networks are trained with left and right oculars, respectively. In the cases of DeepIrisNet-A, DeepIrisNet-B, and Multi-abstract fusion CNN, we tried our best effort to implement these networks from scratch by following \cite{Gangwar2016} and \cite{Soleymani2018}, respectively, as the networks are not publicly available.

\subsection{Experimental results}
This section presents the experimental results on the tasks of periocular recognition. We conducted the experiments on periocular recognition in the wild and controlled environments. We evaluated the performance using Cumulative Matching Characteristic (CMC) curve with 95\% confidence interval (CI).

\vspace{-10pt}
\subsubsection{Performance evaluation on proposed network vs single-stream CNN}
\hspace{8pt} This section analyses the robustness and performance of our proposed network. Table \ref{tab:tab2} presents the performance analysis on the capability of a single-stream CNN with respective features, followed by dual-stream CNN \textit{without} sharing the weights in \textit{conv} and \textit{fc} layers, and our proposed network. Note that, the dual-stream CNN without sharing the weights also implemented the late-fusion and score fusion. This experiment was conducted using Ethnic-ocular dataset.

Table \ref{tab:tab2} shows that our proposed network achieved the highest rank-1 recognition accuracy with 85.0$\pm$1.9\%. However, single-stream CNN only achieved 80.8$\pm$1.4\% and 66.6$\pm$2.2\% with RGB data and OC-LBCP, respectively. As compared to single-stream CNN, the results indicate that the late-fusion layers are significant to correlate the RGB data and OC-LBCP in order to achieve better recognition performance. The analysis demonstrated that our proposed network provides more complementary information than single-stream CNN.

Compared with dual-stream CNN without shared weights, the experimental results in Table \ref{tab:tab2} show that our proposed network is well-performed than dual-stream CNN without shared weights at least 2.9\% improvement. This is because our proposed network utilised the shared \textit{conv} and the fusion \textit{fc} layers to aggregate the RGB data and OC-LBCP. As a result, the proposed have successfully transformed the new knowledge representations in the network to perform better recognition.

\begin{table}[t]
\small
\centering
\caption{Performance analysis on rank-1 recognition accuracy. The highest accuracy is written in bold.}
\label{tab:tab2}
\begin{tabular}{lc}
\hline
\multicolumn{1}{c}{\textbf{Approaches}} & \multicolumn{1}{c}{\textbf{Accuracy (\%)}} \\ \hline
CNN using RGB data & 80.8$\pm$1.4  \\
CNN using OC-LBCP & 66.6$\pm$2.2  \\
Dual-stream CNN (without shared weights) & 82.1$\pm$1.6  \\
\textbf{Proposed network} & \textbf{85.0$\pm$1.9} \\ \hline
\end{tabular}
\vspace{-12pt}
\end{table}

\vspace{-10pt}
\subsubsection{Performance evaluation on proposed network vs benchmark approaches}

\begin{figure}[!t]
\vspace{-10pt}
\centering
\subfigure[UBIPr dataset]{\includegraphics[width=0.36\textwidth]{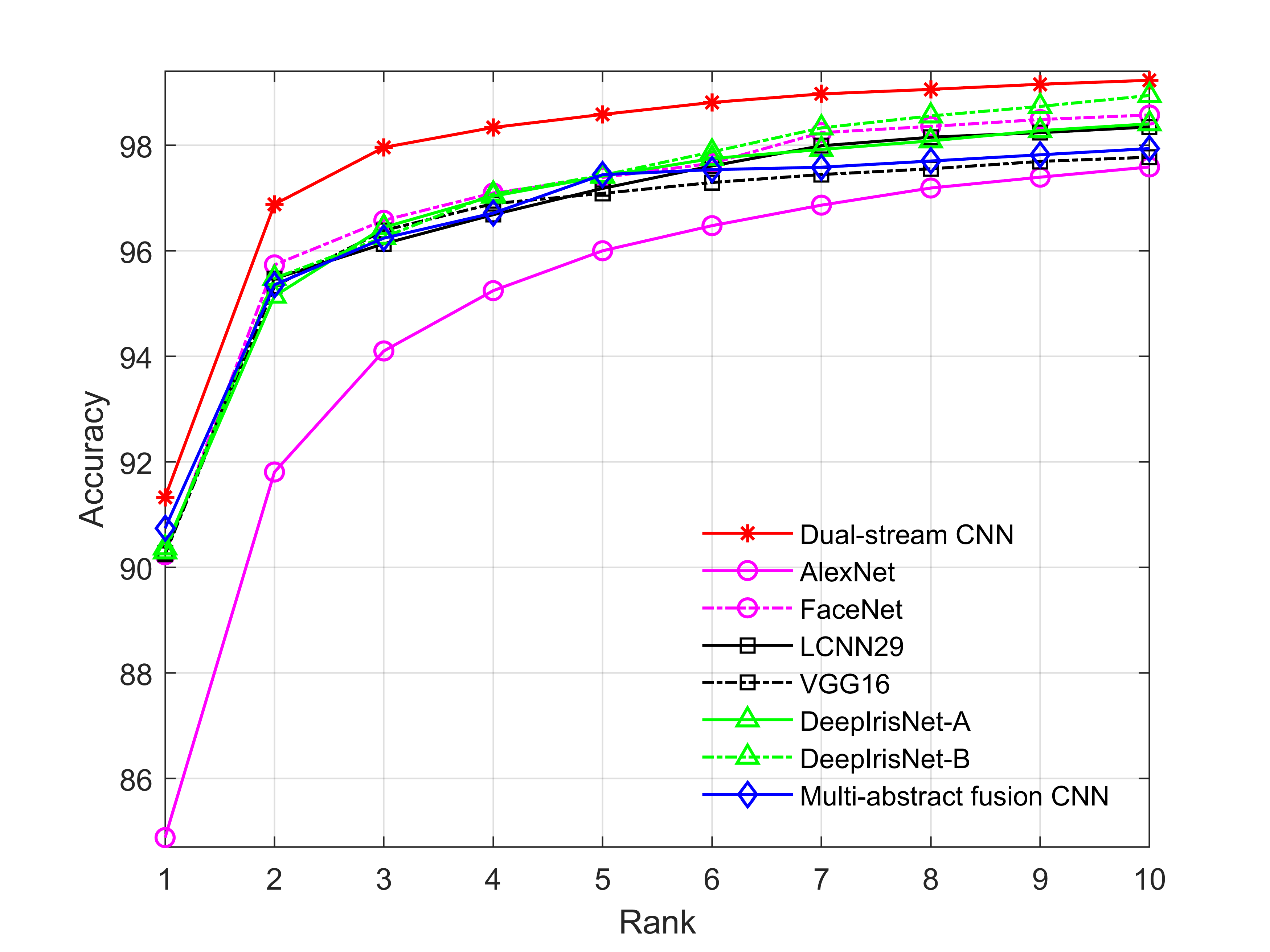}%
\label{fig3b}}\vspace{-10pt}
\subfigure[Ethnic-ocular dataset]{\includegraphics[width=0.36\textwidth]{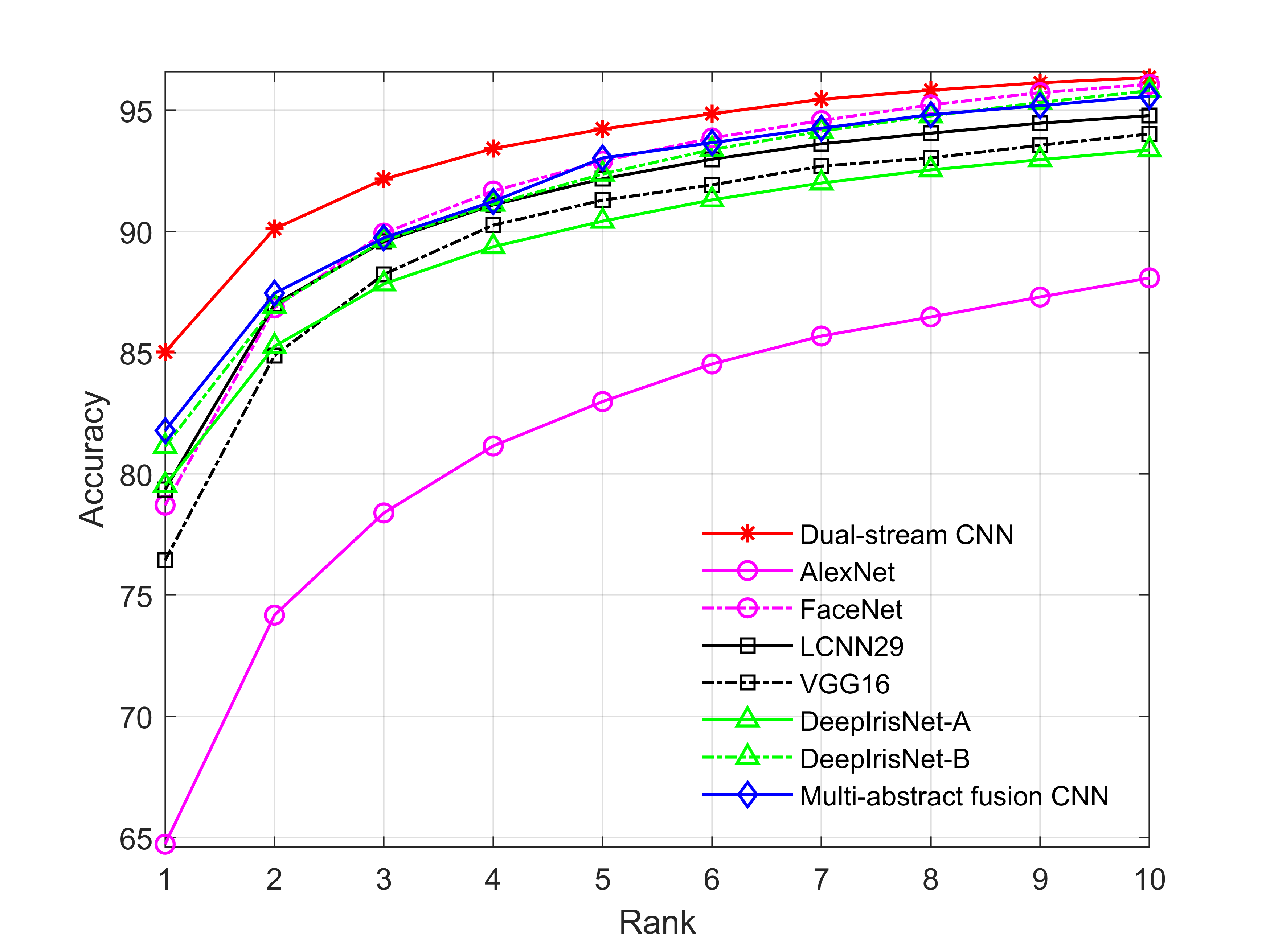}%
\label{fig3c}}
\caption{Performances of CMC curve.}
\vspace{-10pt}
\label{fig:fig4}
\end{figure}


\begin{table*}[t]
\small
\centering
\caption{Evaluation of recognition performance on the Ethnic-ocular dataset and UBIPr dataset. The highest accuracy is written in bold.}
\label{tab:tab3}
\begin{tabular}{p{5.4cm}cccc}
\hline
\multicolumn{1}{c}{\multirow{2}{*}{\textbf{Approach}}} & \multicolumn{2}{c}{\textbf{Ethnic-ocular (\%)}} & \multicolumn{2}{c}{\textbf{UBIPr (\%)}} \\ \cline{2-5} 
\multicolumn{1}{c}{} & \textit{Rank-1} & \textit{Rank-5} & \textit{Rank-1} & \textit{Rank-5} \\ \hline
AlexNet & 64.72$\pm$3.28 & 82.98$\pm$2.52 & 84.88$\pm$2.50 & 96.01$\pm$1.77 \\ 
FaceNet & 78.71$\pm$3.66 & 92.19$\pm$1.59 & 90.24$\pm$1.43 & 97.36$\pm$0.44 \\ 
LCNN-29 & 79.35$\pm$2.64 & 92.17$\pm$1.80 & 90.28$\pm$1.71 & 97.18$\pm$0.67 \\ 
VGG-16 & 76.43$\pm$2.16 & 91.29$\pm$1.54 & 90.24$\pm$1.38 & 97.09$\pm$1.14 \\ 
DeepIrisNet-A & 79.54$\pm$3.12 & 90.43$\pm$2.44 & 90.30$\pm$1.16 & 97.41$\pm$1.07 \\ 
DeepIrisNet-B & 81.13$\pm$3.08 & 92.37$\pm$1.20 & 90.20$\pm$1.66 & 97.43$\pm$0.54 \\ 
Multi-abstract fusion CNN & 81.79$\pm$3.54 & 93.03$\pm$1.33 & 90.75$\pm$1.01 & 97.44$\pm$0.34 \\
\textbf{Proposed network} & \textbf{85.03$\pm$1.88} & \textbf{94.23$\pm$1.26} & \textbf{91.28$\pm$1.18} & \textbf{98.59$\pm$0.44} \\ \hline
\end{tabular}
\vspace{-10pt}
\end{table*}

\paragraph{UBIPr dataset:}
To verify the robustness of our proposed network, we also conducted the performance on more subjective experiment with UBIPr dataset. This dataset consists of 342 subjects with varying different subject-camera distances, poses, illumination, and occlusion. This experiment evaluated the performance of all the approaches with varying pose and subject-camera distances. Six images from each subject were randomly selected as a gallery set; the remaining images were used as a probe set. The selection process was repeated three times.

Table \ref{tab:tab3} presents that dual-stream CNN achieves the highest average rank-1 and rank-5 recognition accuracies with 91.28$\pm$1.18\% and 98.59$\pm$0.44\%, respectively. The second best is achieved by multi-abstract fusion CNN with 90.75$\pm$1.01\% and 97.44$\pm$0.34\% as rank-1 and rank-5 accuracies. Figure 4a shows that our network outperforms most of the benchmark approaches and achieves the highest recall rate against all other approaches for all ranks recognition.
\vspace{-10pt}
\paragraph{Ethnic-ocular dataset:}
We presented the experimental results in Table \ref{tab:tab3} by following the recognition protocol as mentioned in Section 4.2. To evaluate the performance of the proposed network, we compared our results with seven benchmark approaches (see Table \ref{tab:tab3}). For the results of recognition, our proposed network achieved 85.03$\pm$1.88\% and 94.23$\pm$1.26\% as rank-1 and rank-5 accuracies, respectively. Figure 4b illustrates the CMC curve of our proposed network, which outperformed other benchmark approaches from rank-1 to rank-10 recognition accuracies.

Besides, the results proved that our network can learn new features from the late-fusion layers for better recognition. The effectiveness of these fusion layers provides strong support to our assumption that multi-feature learning achieves significantly better results than using raw data.

\section{Conclusion}
This paper outlined a plausible perspective into how machine interpretation of periocular images in the wild could benefit from the RGB data and colour-texture descriptors, known as OC-LBCP. In addition, a dual-stream CNN utilized the late-fusion feature learning, which shown contribute to a more robust feature representation in recognition. We observed that accessing to dual inputs (RGB data and OC-LBCP) significantly outperformed the existing descriptors. We also introduced a new Ethnic-ocular dataset, which consists of a large collection of periocular images based on different ethnic groups for recognising individuals. Good performances were obtained for both controlled and in the wild environments of periocular recognition with the proposed network. However, this work is limited to case of individual who is wearing sunglass. In future, we aim to explore generative adversarial network to reconstruct new periocular images without sunglasses.

{\small
\bibliographystyle{ieee}
\bibliography{mybib}
}

\end{document}